*Review Article*

# Computational Intelligence in Sports: A Systematic Literature Review

**Robson P. Bonidia**  **,[1] Luiz A. L. Rodrigues,[2] Anderson P. Avila-Santos,[3] Danilo S. Sanches,[1] and Jacques D. Brancher[2]**

[1]*Bioinformatics Graduate Program, Federal University of Technology-CP (UTFPR), Paraná, Brazil*
[2]*Computer Science Department, State University of Londrina (UEL), Londrina, Paraná, Brazil*
[3]*Technology College, National Service Industrial Learning of Paraná (SENAI), Londrina, Paraná, Brazil*

Correspondence should be addressed to Robson P. Bonidia; robservidor@gmail.com

Recently, data mining studies are being successfully conducted to estimate several parameters in a variety of domains. Data mining techniques have attracted the attention of the information industry and society as a whole, due to a large amount of data and the imminent need to turn it into useful knowledge. However, the effective use of data in some areas is still under development, as is the case in sports, which in recent years, has presented a slight growth; consequently, many sports organizations have begun to see that there is a wealth of unexplored knowledge in the data extracted by them. Therefore, this article presents a systematic review of sports data mining. Regarding years 2010 to 2018, 31 types of research were found in this topic. Based on these studies, we present the current panorama, themes, the database used, proposals, algorithms, and research opportunities. Our findings provide a better understanding of the sports data mining potentials, besides motivating the scientific community to explore this timely and interesting topic.

## 1. Introduction

The advent of computing has produced a society that feeds on information. Most of the information is in its raw form, known as data [1]. Data are collected and accumulated at an increasing rate [2]. The government, corporate, and scientific areas have promoted significant growth in their databases, overcoming the usual ability to interpret and analyze data, thus, generating the need for new tools and techniques for automatic and intelligent evaluation [2–4].

Therefore, the data mining technique is one of the most competent alternatives to help in the knowledge extraction from large data volumes, discovering patterns, and generating rules for predicting and comparing data, which can help institutions in decision-making and achieve a greater degree of confidence [5, 6].

Historically, the technique of finding useful patterns in data has been named with a variety of names, including data mining, knowledge extraction, identification information, data archeology, and data processing standard [3]. Basically, this technique is a process of exploring significant correlations and trends and discovering intriguing and innovative patterns, as well as descriptive models and comprehensible and predictive data, using statistical and mathematical techniques [4, 7].

Data mining is used to generate knowledge in many scientific, industrial, and mainly business sectors [3]. Moreover, it has been used mostly by statisticians, data analysts, and management information systems. This is because information is the most relevant asset for these organizations, becoming fundamental to gain competitiveness among small, medium, and large companies [8]. Nevertheless, the effective use of data in some areas has gradually developed, as is the case with sports, known for a large amount of information collected from each player, training session, team, games, and seasons [1, 9, 10].

Hence, connoisseurs and experts have dedicated to predict and discuss sporting results. With a large amount of data available (especially since the advent of the Internet), it was natural for statisticians and computer scientists to show interest in discovering patterns and making predictions using these data. The processing of sports data with data mining techniques can not only reduce manual workload and errors, but also improve fairness and development of sports games, assisting coaches and managers in predicting results, assessing players' performance, identifying talents, sporting strategy, and mainly making decision [11].

Based on these assignments, this article aims to perform a Systematic Literature Review (SLR), with the purpose of identifying researches using data mining in the sports field and describes the techniques and algorithms applied and possible research opportunities. Moreover, the current panorama of research, temporal distribution, themes, databases used, and proposals of these works will be presented.

This paper is organized as follows: Section 2 discusses the systematic review methods. Section 3 reports research planning. Sections 4 and 5 present how the SLR was conducted and the results obtained. Finally, Section 6, described the final considerations.

## 2. Systematic Review Methods

This study used the systematic review method which , according to Brereton et al. [12], allows a rigorous and reliable evaluation of the research carried out within a specific topic. This type of investigation presents a summary of evidence through comprehensible systematic search methods and synthesis of selected information [13]. Moreover, this methodology has been widely used [12–15]. Thus, we based our research on previous works recommendations, which is to divide this process into three stages: planning, conducting, and analysis of results. Therefore, the following sections describe how we approached each one of these stages.

## 3. Research Planning

The planning stage addressed the scientific questions definition, the intervention of interest specification, databases identification, keywords definition, search strategies, criteria for inclusion and exclusion, and articles quality [12, 16]. Therefore, the following issues were defined as research questions (RQn):

(i) RQ1: What is the current researches overview?

(ii) RQ2: What are the most used techniques?

(iii) RQ3: What is the temporal distribution of the works?

(iv) RQ4: What are the most cited research papers?

(v) RQ5: What are the datasets?

(vi) RQ6: What was the result of the researches?

(vii) RQ7: What are the algorithms and methods applied?

(viii) RQ8: What are the research opportunities?

Usually, inclusion, exclusion, and quality criteria are determined after the definition of research questions [17]. Hence, Table 1 establishes the criteria used in this study.

TABLE 1: Inclusion, exclusion, and quality criteria.

| | Criteria |
|---|---|
| | Studies in English |
| Inclusion | Article, Conference or Methodology paper |
| | Studies relevant to the sports data mining techniques |
| | Researches that do not accomplish the quality criteria |
| Exclusion | Studies outside the context of work |
| | Studies written in another language than English |
| | Studies published before 2010 |
| Quality | Studies with full results |
| | Studies with different proposals/results |

Aspects of this process may include decisions about the type of revisions that should be included in the research, which is used to manage the selection criteria in a subset of primary studies [16, 18]. Therefore, in order to ensure the quality of this review, every article found will be analyzed according to Title, Abstract, Keywords, Proposed Mechanism, Results, and Conclusion. Furthermore, this article used the following electronic databases to find papers:

(i) *ACM Digital Library*: dl.acm.org

(ii) *IEEE Xplore Digital Library*: ieeexplore.ieee.org

(iii) *Science Direct*: www.sciencedirect.com

(iv) *Semantic Scholar*: www.semanticscholar.org

Finally, the selected method to search in these databases was boolean recovery. Essentially, it divides a search space, identifying a subset of documents in a collection, according to the criteria of consultation [19]. In our case, the key is the following string: *(("sports" OR "sport" OR "sports science") AND ("data mining" OR "mining" OR "computational intelligence" OR "machine learning" OR "deep learning" OR "artificial intelligence"))*.

## 4. Execution Plan

This stage involves five steps: (1) performing the search in the selected databases; (2) comparison of searches results to exclude repeated papers; (3) application of inclusion, exclusion, and quality criteria; (4) evaluation of all studies that passed the initial review; and (5) data synthesis [12, 16].

Figure 1 displays this flow for our SLR. Basically, the first phase consisted of executing the search strings in all databases, which found a large set of 1582. Thus, to aid the review and achieve better accuracy and reliability, the StArt tool (State of the Art through Systematic Reviews) was used. This tool has the purpose of supporting researchers in their systematic analysis [20–22].

It uses BibTeX extensions (bibliographic formatting file used in LaTeX documents) to perform these analyses. Therefore, these extensions were extracted from the aforementioned databases as well. It is important to note that the BibTeX files were exported without any filter, which explains the number of researches returned.

Thereafter, works published before 2010 were eliminated, returning an amount of 1172 titles (410 rejected articles).

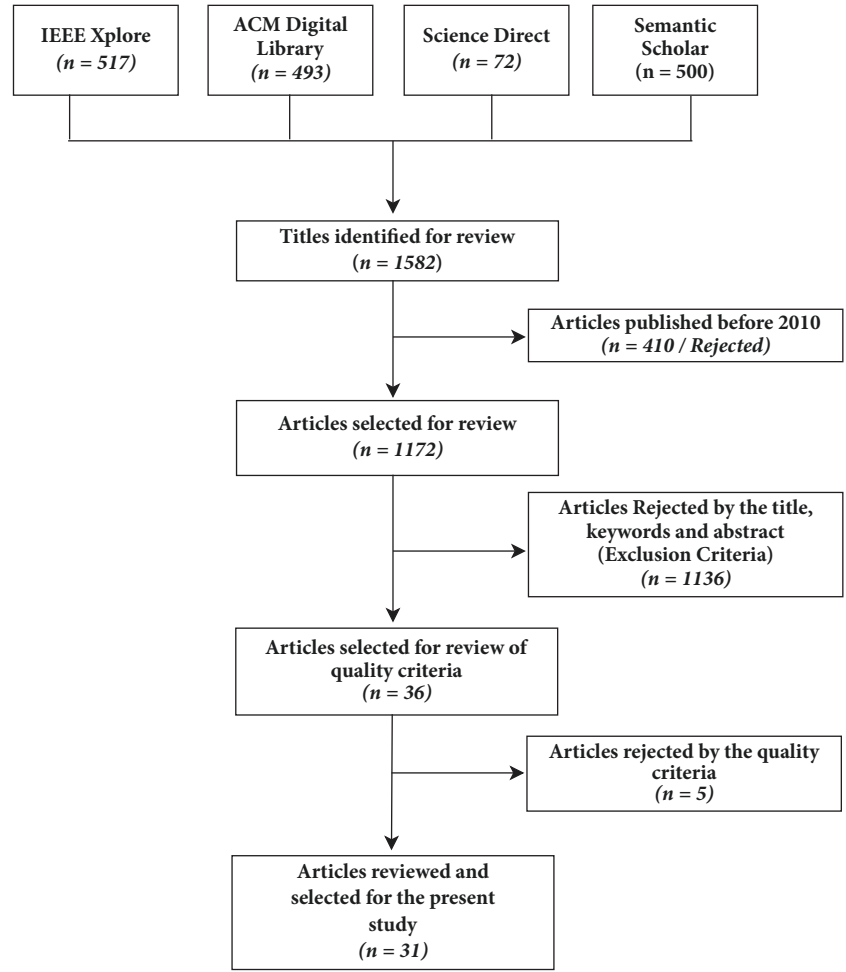

Figure 1: Systematic review research flow.

Nevertheless, in order to refine the search and eliminate articles that were outside the scope of this review, a careful analysis was applied in the titles, keywords, and abstracts, according to the exclusion criteria (see Table 1). It eliminated 1136 works, remaining a preselection set of 36 titles for quality analysis. It is important to emphasize which queries using keywords return several titles that are far from the context of this article; this justifies the amount of rejected research.

Finally, after the works preselection, a synthesis of the data was performed, with the objective of applying an evaluation based on the stated quality criteria. Thereby, of the 36 articles, 5 were eliminated (they have not demonstrated the methods or techniques applied), leading to a final set of 31 articles with relevant information about data mining in sports.

## 5. Results

In this section, the results of this SLR are presented. Thus, each subsequent subsection will answer the issues raised at the beginning of the research.

*5.1. RQI: Current Researches Overview.* The works selected by this SLR are presented in Table 2. Examining them, we will provide a report regarding the current panorama of sports data mining.

Therefore, to provide an overview of the thematic types that have been proposed in the articles, they were categorized in nine classes: Motion Analysis; Performance Evaluation; Sports Data Capture; Generating Eating Plans; Training Planning; Strategic Planning; Predicting Results/Patterns; Sports Data Analytics; and Decision-Making Support. For a better understanding of these categories Table 3 introduces them.

(i) **Motion Analysis**: themes focused on strategies for the recognition and classification of patterns and skills through the motion analysis.

(ii) **Performance Evaluation**: guide coaches or athletes on how to improve performance and eliminate errors through data mining.

(iii) **Sports Data Capture**: category presenting the papers that developed or studied sports information extraction systems.

(iv) **Generating Eating Plans**: articles reporting research on the automatic generation of optimal food plans for athletes through artificial intelligence.

Table 2: Selected papers (IDs and references).

| ID | Reference | ID | Reference |
|----|-----------|----|-----------|
| 1  | [23]      | 17 | [24]      |
| 2  | [25]      | 18 | [26]      |
| 3  | [27]      | 19 | [28]      |
| 4  | [29]      | 20 | [30]      |
| 5  | [31]      | 21 | [32]      |
| 6  | [33]      | 22 | [34]      |
| 7  | [35]      | 23 | [36]      |
| 8  | [37]      | 24 | [38]      |
| 9  | [39]      | 25 | [40]      |
| 10 | [41]      | 26 | [42]      |
| 11 | [43]      | 27 | [44]      |
| 12 | [45]      | 28 | [46]      |
| 13 | [47]      | 29 | [48]      |
| 14 | [49]      | 30 | [50]      |
| 15 | [51]      | 31 | [52]      |
| 16 | [53]      |    |           |

Table 3: Thematic types.

| Theme | Articles |
|-------|----------|
| Motion Analysis | [23, 27] |
| Performance Evaluation | [31, 33] |
| Sports Data Capture | [35] |
| Generating Eating Plans | [37] |
| Training Planning | [39, 41, 43, 45, 47, 49] |
| Strategic Planning | [24, 51, 53] |
| Predicting Results/Patterns | [26, 28, 30, 32, 34, 36, 38, 40, 42] |
| Sports Data Analytics | [25, 29, 44] |
| Decision Making Support | [46, 48, 50, 52] |

(v) **Training Planning**: intelligent studies for planning sports training sessions, techniques to support the definition of training sessions and quick feedback systems.

(vi) **Strategic Planning**: researches developed from systems to support strategic planning in sports using data mining modules, tactical analysis techniques, and computational intelligence.

(vii) **Predicting Results/Patterns**: topics predicting game results in sports such as college football, elite soccer, basketball, and golf.

(viii) **Sports Data Analytics**: data mining platforms that aid the diagnosis of sports records.

(ix) **Decision-Making Support**: articles that investigated the potential of data mining or computational intelligence with the purpose of support in the decision-making process.

Papers distribution between the aforementioned categories is exposed in Table 3. As can be seen, the growth of sports data mining, even if limited, presents relevant works aiming to automate empirical tasks and improve the reliability of strategies, predictions, and training decisions. Moreover, it is noted that the category Predicting Results/Patterns presents the greatest amount of works (9), followed by Training Planning (6), Decision-Making Support (4), Strategic Planning and Sports Data Analytics (3), Motion Analysis and Performance Evaluation (2), and finally Sports Data Capture and Generating Eating Plans (1).

*5.2. RQ2: Most Used Techniques.* A words cloud is demonstrated in Figure 2, where the size of each word reflects its frequency of occurrence. This cloud is based on words contained in the titles of the selected articles.

The preponderance of words suggests that these are the most used techniques to create computational intelligent systems in sport's field. Thus, as can be seen in the figure, data mining and machine learning are the most used, considering that words related to these techniques often appear in the papers, based on their size in this figure.

*5.3. RQ3: Papers Temporal Distribution.* Analyzing the temporal distribution of included articles, it was noticed that years 2014 and 2016 reported the highest amount of publications, with 12 articles (6 each year), representing 38.71% of the works reviewed. The years 2010, 2013, and 2017 together presented the same total number of articles (12-38.71%), but with only 4 works in each year. The year of 2012 had 3 (9.68%) articles, whereas 2015 and 2018 had only 2 (12.90%). This distribution is shown in Figure 3.

*5.4. RQ4: Most Cited Researches.* In total, the top 5 articles contributed 133 citations related to sports data mining, as can be seen in Table 4. The research of Novatchkov [43] published in Journal of sports science (2013) was the most cited (36 times). The remaining papers were published in Intelligent Systems and Informatics ([42], 31 times), Computational and Business Intelligence ([47], 26 times), Procedia Computer Science ([40], 23 times), and IFAC Proceedings Volumes ([27], 14 times).

*5.5. RQ5/RQ6: Analyzed Datasets.* This section presents the proposals or results as well as datasets used by the selected works in Tables 5 and 6, respectively. We highlight that data mining depends on the quality and quantity of data in order that algorithms yield satisfactory results. Furthermore, another limitation is to find a safe source and select relevant attributes to investigate the problem. Therefore, it is relevant to SLR report datasets used in the literature.

It is observed in Table 5 that the great majority seeks to find patterns to predict results, strategies, training sessions, and mainly support in decision-making. Other papers discuss the motion analysis, generating eating plans and data extraction tools. Nonetheless, the SLR demonstrated that the last 3 citation topics feature little-explored sports field. Furthermore, a great need that we perceive during the review is the data extraction. According to [35], text data related to sports are semistructured or even unstructured. Essentially, these texts contain relevant information and are available on the web; however, extracting data and useful knowledge of these documents adequately is difficult and costly. This is confirmed in the datasets exposed in Table 6, which reflect the lack of model data to work with sports data mining.

FIGURE 2: Frequently occurring words in articles under review.

FIGURE 3: Work's temporal distribution.

TABLE 4: Top 5 articles according to citation frequency.

| Article | Journal/Conference | Year | Citation |
| --- | --- | --- | --- |
| [43] | Journal of sports science & medicine | 2013 | 39 |
| [42] | Intelligent Systems and Informatics | 2010 | 31 |
| [47] | Computational and Business Intelligence | 2013 | 26 |
| [40] | Procedia Computer Science | 2014 | 23 |
| [27] | IFAC Proceedings Volumes | 2012 | 14 |

*5.6. Data Mining Techniques.* Here, we present a brief description of the data mining techniques used by the surveyed papers in this SLR. Five main techniques were identified: classification, clustering, association, regression, and heuristics.

  (i) **Classification:** consists of developing a model that can be applied in the prediction of events with a predefined number of outcomes (e.g., is the player's arm moving? Will my team win this game?). Fundamentally, this technique has a wide range of applications, including medical diagnosis, fraud detection, credit approval, target marketing, and prediction of sports results [54].

  (ii) **Clustering**: analyzes data to find groups of items that are similar to each other according to some similarity metric. Essentially, this technique is used for unsupervised learning, when we do not know the answer to the investigated problem (e.g., finding players with similar skills or team using similar tactics without a direct indicator of these characteristics) [55].

  (iii) **Association**: consists of identifying which attributes are related, extracting associations between a large set of items or events (e.g., players who are good in one task are also good in another) [7, 54, 56].

  (iv) **Regression**: determines the relationship between a dependent variable (target field) and one or more independent variables. The dependent variable is the one whose values you want to predict, while the independent ones are the variables based on the prediction. In this case, the predicted values are continuous (e.g., how old is this player? How many points will my team score?) [7].

  (v) **Heuristic**: the heuristic term means to discover. This technique is inspired by intuitive processes, whose purpose is to find a good solution at an acceptable computational cost. Fundamentally, heuristic algorithms seek viable solutions to optimization problems in which the complexity and time available for their solution do not allow the exact result [57, 58].

Furthermore, Table 7 shows, for each theme, the techniques that have been employed to address it. As can be seen, the classification was the most used (51.28%), followed by clustering, heuristics, and association with 17.95%, 15.39%, and 12.82% applications, respectively, while that the regression was adopted by a single research (2.56%).

*5.7. RQ7: Applied Algorithms by the Articles.* This section aims to explore the algorithms used in the revised works, as in Table 8 (Table follows as order the thematic types), in addition to making a brief introduction to the most used algorithms.

  (i) **Bat algorithm**: this algorithm was developed in 2010 by Yang. The Echolocation process that bats use during their flight was the inspiration. Besides being used for orientation this mechanism helps the bats to find their prey. Essentially, the bat algorithm is based on the population of individuals where each one represents the candidate solution [49].

  (ii) **Differential Evolution**: this algorithm was proposed in 1995 by Storn and Price [59]. The DE is a technique of the stochastic research efficient and powerful. It was based on population to solve the problem of continuous space optimization that has been largely applied in the most several scientific fields [60].

Table 5: Proposals/results of the works.

| Paper | Proposal/Result |
|-------|-----------------|
| [27] | Research of strategies most used for the recognition and classification of human movement patterns. |
| [23] | Analysis of sports skills data with temporal series image data retrieved from films focused on table tennis. |
| [33] | Guide the athletes on how to improve their performance and how to eliminate errors related to the selection of the proper running strategy through the differential evolution algorithm. |
| [31] | Proposes a new clustering algorithm based on ant colony optimization. |
| [35] | Proposed the development of an information extraction system wherein its purpose was to obtain data frames of multiple sports performance documents. |
| [37] | Automatic generation of optimal food plans for athletes, through the particle swarm optimization algorithm. |
| [47] | Proposed an automated personal trainer. |
| [49] | Solution for automatic planning of training sessions. |
| [39] | A new solution capable of adapting training plans. |
| [41] | A framework to automatically analyze the physiological signals monitored during a test session. |
| [43] | Implementation of artificial intelligence routines for automatic evaluation of exercises in weight training. |
| [45] | Presented three geometric/temporal features of pen trajectories used in a cognitive skills training application for elite basketball players. |
| [24] | An data mining algorithm to soccer tactics using association rules mining. |
| [53] | Discussed the application of the association rule mining in sports management, especially, in cricket. |
| [51] | Presented a relational-learning based approach for discovering strategies in volleyball matches based on optical tracking data. |
| [36] | A generalized predictive model for predicting the results of the English Premier League. |
| [30] | A data analysis to identify important aspects separating skilled golfers from poor. |
| [38] | Compared the performance of algebraic methods to some machine learning approaches, particularly in the field of match prediction. |
| [40] | A sports data mining approach, which helps discover interesting knowledge and predict results from sports games such as college football. |
| [42] | Data mining techniques for predicting basketball results in the NBA (National Basketball Association). |
| [28] | Developed a tool COP (Cricket Outcome Predictor), which outputs the win/loss probability of a match. |
| [32] | Classify players into regular or All-Star players from the National Basketball Association and identify the most important features that make an All-Star player. |
| [26] | Designed and built a big data analytics framework for sports behavior mining and personalized health services. |
| [34] | Provides a prediction model of sports results based on knowledge discovery in database. |
| [25] | A machine learning system with unsupervised learning and supervised learning components to analyze chess data. |
| [44] | Concluded that the most important elements in basketball are two-point shots under the arch and defensive rebound. |
| [29] | A data mining approach for classification and identification of golf swing from weight shift data. |
| [52] | Describes machine learning techniques that assist cycling experts in the decision-making processes for athlete selection and strategic planning. |
| [46] | Predict match outcomes in the 2015 Rugby World Cup. |
| [48] | Presented a visualization system that uses statistics and movement analysis. Basically, the type of pattern of attack and play can be understood dynamically and visually. |
| [50] | Conducted a study on a decision support system for techniques and tactics in sports. |

(iii) **Particle Swarm Optimization**: it is a bioinspired computational algorithm in the social behavior metaphor about the interaction between individuals (particles) of a group (swarm), developed in 1995 by Kennedy and Eberhart. This algorithm was implemented based on the observation of flocks of birds and shoals of fish in search of food in a certain region [61–63].

(iv) **Ant Colony Optimization**: the Ant Colony Optimization (ACO) is a bioinspired algorithm by the foraging behavior of some species of ants. Essentially, this technique applies the pheromone method that the ants deposit on the ground to demarcate a more favorable path and that must be followed by other members of the colony. The ACO uses a similar method to solve optimization problems [64, 65].

TABLE 6: Datasets.

| Paper | Dataset |
|---|---|
| [27] | - |
| [23] | Moving images of 15 male college tennis players. |
| [33] | Data of Laguna Poreč half-marathon (2017) and Ormož half-marathon (2017). |
| [31] | A set of sports performance data of college students. |
| [35] | - |
| [37] | Training plan generated by an Artificial Sports Trainer and a list of potential nutrition. |
| [47] | - |
| [49] | Used exercise datasets for training. |
| [39] | Sports training plans generated by an Artificial Sport Trainer. |
| [41] | Training data and competitions of a cycling mode athlete. |
| [43] | Used sensors attached to various exercise equipment, allowing the collection of characteristics during the workout. |
| [45] | - |
| [24] | A football match data from European Cup 2008's final match - Spain vs Germany. |
| [53] | Data of matches played by India. |
| [51] | The data from the FIVB Volleyball World Championships finals that were held in Poland and Italy in 2014. |
| [36] | Data from 2005 to 2016, spanning 11 seasons of the English Premier League. |
| [30] | Data from 275 male golfers. |
| [38] | Website data http://www.football-data.co.uk/. |
| [40] | Real-life statistical data from cfbstats.com for past college football games. |
| [42] | A dataset comprising 778 games from the regular part of the 2009/2010 NBA season. |
| [28] | Data from One Day International (DOI) matches during the time period 2001-2015 for each team - http://www.cricinfo.com |
| [32] | An NBA men basketball dataset that is publicly available at open source sports in the period 1937 till 2011. |
| [26] | - |
| [34] | - |
| [25] | Data from 500 games from each of the 10 grandmaster chess players, a total of 5000 chess games. |
| [44] | Data from the First B basketball league for men in Serbia, from seasons 2005/06, 2006/07, 2007/08, 2008/09 and 2009/2010. |
| [29] | Weight shift data from golf experiments conducted by the research team. |
| [52] | Competition results for senior riders including the Australian Championships 2009, World Championships 2007–2010, UCI World Cup Melbourne 2010, UCI World Cup Cali 2011, UCI World Cup Beijing 2011, UCI World Cup Manchester 2011 and Oceania Championships 2010. |
| [46] | History of statistical data, ranking, and points of 20 rugby teams. |
| [48] | American football game data. |
| [50] | - |

TABLE 7: Classification of thematic types by techniques.

| Theme | Classification | Clustering | Association | Regression | Heuristic |
|---|---|---|---|---|---|
| *Motion Analysis* | [23, 27] | - | - | - | - |
| *Performance Evaluation* | - | [31] | - | - | [31, 33] |
| *Sports Data Capture* | [35] | - | - | - | - |
| *Generating Eating Plans* | - | - | - | - | [37] |
| *Training Planning* | [43, 47] | [41, 47] | [45] | - | [39, 49] |
| *Strategic Planning* | [51] | - | [24, 53] | - | - |
| *Predicting Results/Patterns* | [28, 30, 32, 34, 36, 38, 40, 42] | [26] | [34] | - | - |
| *Sports Data Analytics* | [25, 29, 44] | [25] | - | - | [29] |
| *Decision Making Support* | [46, 48, 52] | [50, 52] | [50] | [50] | - |

TABLE 8: Algorithms.

| Paper | Algorithm |
|---|---|
| [27] | Artificial Neural Networks, Statistical Classifiers and Hidden Markov Models. |
| [23] | C4.5, Random Forest and Native Bayes Tree. |
| [33] | Differential Evolution. |
| [31] | k-means, Ant Colony Optimization. |
| [35] | Naive Bayes. |
| [37] | Particle Swarm Optimization. |
| [47] | - |
| [49] | Bat Algorithm. |
| [39] | Particle Swarm Optimization. |
| [41] | K-Means. |
| [43] | Artificial Neural Networks. |
| [45] | AISReact. |
| [24] | Association Rule Mining Algorithms. |
| [53] | Association Rule Mining Algorithms. |
| [51] | Inductive Logic Programming |
| [36] | Gaussian Naive Bayes, Support Vector Machine and Random Forest. |
| [30] | Random Forest and Classification and Regression Trees. |
| [38] | Linear Algebra Methods, Artificial Neural Networks and Random Forest. |
| [40] | Decision tree, Artificial Neural Networks and Support Vector Machine. |
| [42] | Naive Bayes, Decision tree, Support Vector Machine and K Nearest Neighbors. |
| [28] | Naive Bayes, Support Vector Machine and Random Forest. |
| [32] | Random Forest. |
| [26] | K-means. |
| [34] | Artificial Neural Networks. |
| [25] | Hierarchical Clustering. |
| [44] | Artificial Neural Networks. |
| [29] | Particle Swarm Optimization, Support Vector Machine, C4.5. |
| [52] | Bayesian Belief Networks, Naive Bayes and K-means. |
| [46] | Random Forest. |
| [48] | - |
| [50] | - |

(v) **Artificial Neural Networks:** Artificial Neural Networks (ANN) are widely used in machine learning. Fundamentally, this brain-inspired tool is meant to replicate the way humans learn. ANNs are excellent tools for finding patterns that are complex or numerous [66, 67].

(vi) **Support Vector Machine**: the SVM algorithms are based on the theory of statistical learning, developed by Vapnik [68] from studies initiated in [69]. This method establishes some principles that must be followed in obtaining classifiers with good generalization, defined as their ability to correctly predict the class of new data from the same domain in which learning occurred.

(vii) **CART Algorithm**: the CART algorithm is based on classification and regression trees. Basically, the algorithm builds a binary decision tree that divides a node into two child nodes repeatedly, starting with the root node that holds all the learning samples. The CART algorithm search all possible variables and values with the goals to detect the best one [70].

(viii) **Neural Network Ensemble**: it is a kind of modular neural network encompassing a set of whole problem classifiers where individual decisions are combined to classify new examples. It is a paradigm of learning where many neural networks are used together to solve a problem. This demonstrates that the generalization capacity of a neural network can be significantly improved through the use of several neural networks [71].

(ix) **Bayesian Classifier:** Bayesian classifiers use the Bayes theorem to perform probabilistic classification. They are statistical classifiers that classify an object into a given class based on the probability that the element belongs to this class. This theorem produces results quickly. It has great accuracy when applied on a big data if compared to the results produced by decision trees and neural networks. Then it is defined by the formula $P(H \mid E) = P(E \mid H)P(H)/P(E)$, where $P(H \mid E)$ is the probabilistic of some hypotheses $H$ which are true whereas the evidence $E$ is noticed. $P(H)$ is the probability that the hypothesis is true regardless of the evidence and $P(E)$ is the prior probability that the evidence will be observed [72].

(x) **Random Forest**: it was proposed in 2001 by Breiman with the goals to combine the classifiers created for several decision trees. This classifier is computationally efficient when each tree is built individually. Therefore, it is a combination of predictive trees which one depends on the vector of random values. This vector has the same distribution for all of the trees in the forest [73, 74].

(xi) **C4.5 Algorithm**: this algorithm builds a decision tree with a division and conquest strategy. Each node in the tree is associated with a set of cases. Furthermore, the cases are weights assigned to take into account the values of unknown attributes. The decision trees generated by it can be used for classification and are known as statistical classifiers [75].

(xii) **K-means Algorithm** this is a simple unsupervised learning algorithm that solves the known clustering problem. The procedure follows a simple way of classifying a given set of data across a number of groups [76, 77].

(xiii) **Hierarchical Clustering**: it is an approach of clustering analysis which has the purpose of building a hierarchy of groups. Strategies for hierarchical grouping are usually divided into two types. Bottom-up: in this method, each observation is assigned to its own group. Then, the similarity between each of the agglomerates is calculated and joined together the two most similar agglomerates. These procedures are repeated until there is only one group. Top-down: in this method, all observations are assigned to a single group; after that the group is divided into two less similar ones. Finally, it proceeds recursively in each group, until there is one for each observation [78].

It is important to emphasize that the revised works cited other algorithms for sports data mining. However, Table 8 is formed only by the algorithms applied in the surveys. Therefore, some of the algorithms mentioned are as follows:

(i) **A Priori Algorithm**: it is a classic algorithm used for data mining for association and learning rules that were proposed in 1994 by Agrawal and Srikant. This algorithm does a recursive search in the database seeking the frequent sets. It presents three phases: generation of candidate sets, pruning of candidate sets, and support calculation [79].

(ii) **Backpropagation Algorithm**: this algorithm was developed in the 70s; however it was fully appreciated after a famous article about the 80s. This article was written by David Rumelhart, Geoffrey Hinton, and Ronald Williams who described several neural networks in a scenario where backpropagation worked faster than previous approaches, making it possible to use neural networks to solve previously unsolvable problems. This algorithm is a multilayer network that uses an adjustment weight based on sigmoid function, such as delta rule. Backpropagation is a supervised learning method, where the function goals are known [80].

(iii) **Multilayer Perceptron Network**: this algorithm uses the backpropagation learning mechanism that is normally used for pattern recognition. Fundamentally, MLP is a multilayer feedforward network that uses a supervised learning mechanism based on the adjustment of its parameters according to the error between the desired and calculated output [7, 81].

(iv) **Levenberg-Marquardt Algorithm**: the Levenberg-Marquardt (LM) Algorithm is an iterative technique that identifies the minimum of a multivariate function, which is exposed with the sum of squares of real nonlinear functions. It is a standard technique for nonlinear least squares problems widely adopted [82, 83].

*5.8. RQ8: Research Opportunities.* Topics related to sports data mining are relevant. However, this domain still has several branches to be explored. Thus, in Table 9, we mapped the sports modalities or application field in which data mining was applied to papers addressing the topics. The goal of this table is to demonstrate possible gaps and future works to be investigated.

According to Table 9, general applications present 29.03% (9) of the papers. Essentially, this topic refers to surveys that were developed without a fixed scope in any sporting modality or field. Nevertheless, studies selected in the literature used more frequently basketball and football/soccer, covering 12.90% (4) and 9.67% (3), respectively. Table tennis, weight training, cycling, cricket, and golf were only 6.46% (2) each. Running, volleyball, American football, chess, and rugby were addressed in a single work each (3.22%). Thereby, there are distinct gaps to be explored since, considering high-quality works published from 2010 to 2018, few areas have been explored while there are many fields that could benefit from these applications, such as swimming, athletics, hockey, boxing, fencing, and tennis. That is, even with the growth sports data mining, there are still numerous fields of application to be studied. Therefore, seeking to aid researchers that intend to contribute to the community on these topics, a set of open hypothesis that could be examined in future studies are shown:

(i) **Analysis of sports performances**: helping to highlight good and bad techniques, beyond the team performance, through data mining, would assist coaches in decision making.

(ii) **Quick feedback systems**: data mining tools using systems that incorporate feedback sensors, to acquire values of biomechanical, physiological, cognitive, and behavioral parameters of the training, could help to improve the performance of athletes in competition.

TABLE 9: Sports modalities or application field.

| Modalities/Field | Articles |
| --- | --- |
| Table Tennis | [23, 50] |
| Running | [33] |
| Weight Training | [43, 49] |
| Cycling | [41, 52] |
| Basketball | [32, 42, 44, 45] |
| Volleyball | [51] |
| Football/Soccer | [24, 36, 38] |
| Cricket | [28, 53] |
| Golf | [29, 30] |
| American Football | [40] |
| Chess | [25] |
| Rugby | [46] |
| General Application | [26, 27, 31, 34, 35, 37, 39, 47, 48] |

(iii) **Methods for talent detection**: systems capable of crossing information from athletes and find performance and development patterns can assist in detecting potential talents.

(iv) **Artificial Trainers**: smart tools to plan training sessions, specifying requirements, constraints, and goals, can present positive effects on athletes.

Finally, it is important to emphasize that the sports domain presents several others subjects in which data mining can be applied. Nevertheless, the objective of this section was to report themes and to highlight some hypotheses for future studies, based on the reviewed papers.

## 6. Final Considerations

In recent years, sports data mining has evolved. Consequently, many sports organizations have noticed that there is a wealth of unexplored knowledge in the data extracted by them. This is because even a small additional view of the variables can decide several factors, thus increasing the competitive advantage of the teams over their rivals. That is, data mining transfers to sports a higher degree of professionalism and reliability. Therefore, this article covered the last eight years (2010-2018) of papers available in relevant databases. The review issues considered methods, information, and applications. Thereby, the current panorama, temporal distribution, themes, the datasets used, proposals, and results of these revised works were presented. Moreover, techniques, algorithms, methods, and research opportunities have been reported.

As a result, we find 31 articles relevant that were separated into nine thematic types, whose highest frequency of publication was in the years 2014 and 2016. We also present the most cited articles, their datasets, and results. Regarding data mining techniques, the classification was most applied. Finally, possible areas to be explored were reported, such as swimming, athletics, hockey, boxing, fencing, and tennis. It is expected that this article provides an important source of knowledge for future researches, beyond encouraging new studies.

## Conflicts of Interest

The authors declare that there are no conflicts of interest regarding the publication of this paper.

## Acknowledgments

The authors thank the Federal University of Technology, Paraná (UTFPR, Grant: April/2018), and Coordenação de Aperfeiçoamento de Pessoal de Nível Superior, Brasil (CAPES), Finance Code 001, for their financial support.

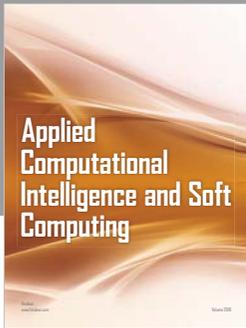
Applied
Computational
Intelligence and Soft
Computing

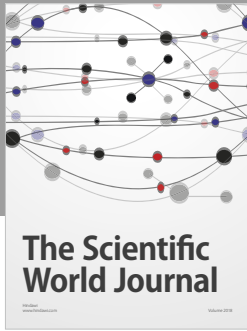
**The Scientific
World Journal**

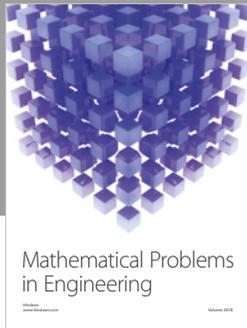
Mathematical Problems
in Engineering

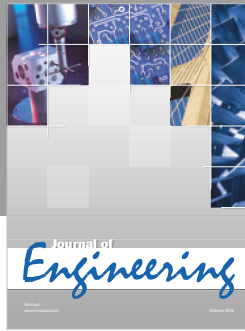
Journal of
*Engineering*

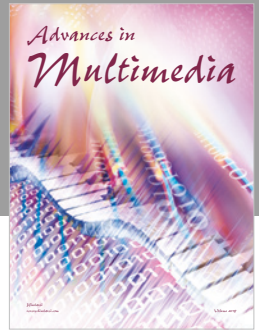
Advances in
*Multimedia*

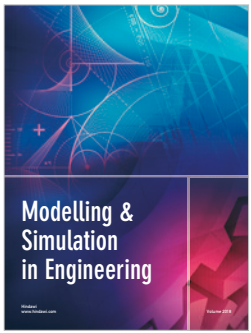
Modelling &
Simulation
in Engineering

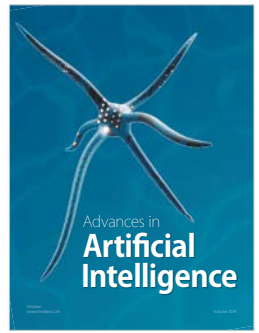
Advances in
**Artificial
Intelligence**

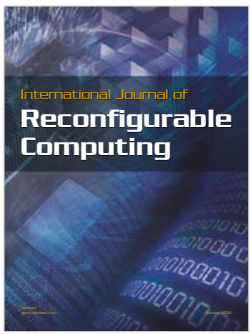
International Journal of
**Reconfigurable
Computing**

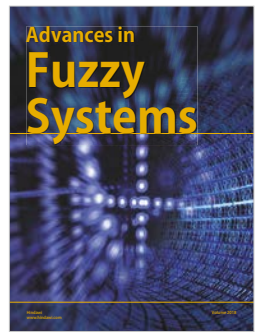
Advances in
**Fuzzy
Systems**

**Hindawi**

Submit your manuscripts at
www.hindawi.com

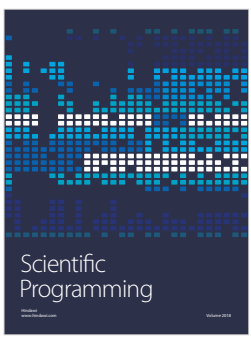
Journal of
**Robotics**

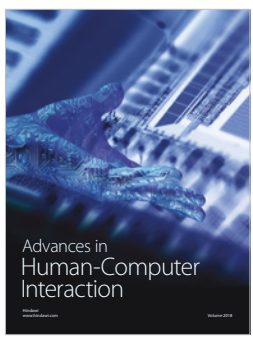
Advances in
**Human-Computer
Interaction**

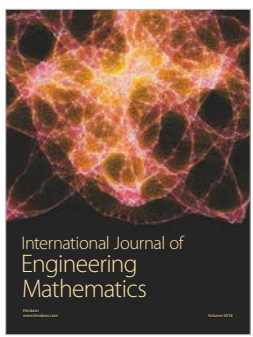
International Journal of
**Engineering
Mathematics**

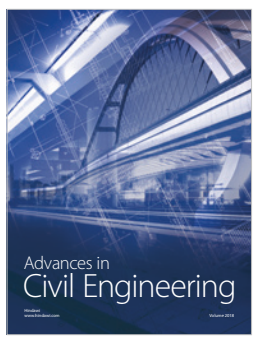
Advances in
**Civil Engineering**

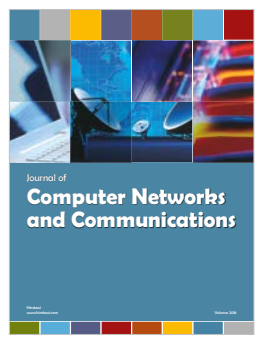
Journal of
**Computer Networks
and Communications**

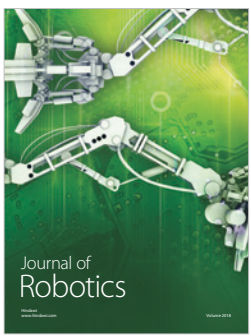
Scientific
Programming

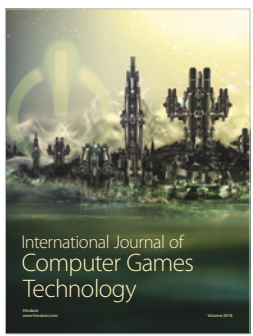
International Journal of
**Computer Games
Technology**

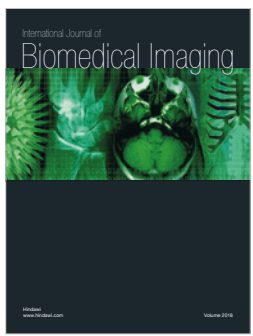
International Journal of
**Biomedical Imaging**

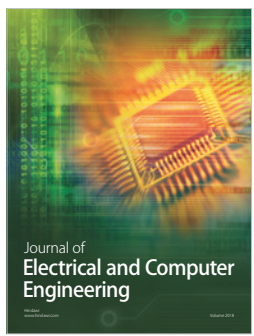
Journal of
**Electrical and Computer
Engineering**

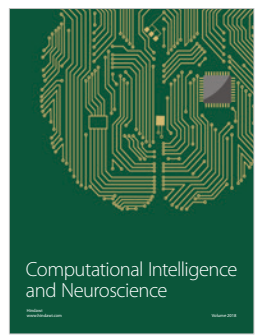
Computational Intelligence
and Neuroscience